# Machine learning approach for segmenting glands in colon histology images using local intensity and texture features

Rupali Khatun
Dept of Computer Applications
Amity University
Kolkata, India
rupali.khatun@live.com

Soumick Chatterjee
Biomedical Magnetic Resonance,
Faculty of Natural Science
Data & Knowledge Engineering Group,
Faculty of Computer Science
Otto von Guericke University
Magdeburg, Germany
soumick.chatterjee@ovgu.de

*Abstract*— Colon Cancer is one of the most common types of cancer. The treatment is planned to depend on the grade or stage of cancer. One of the preconditions for grading of colon cancer is to segment the glandular structures of tissues. Manual segmentation method is very time-consuming, and it leads to life risk for the patients. The principal objective of this project is to assist the pathologist to accurate detection of colon cancer. In this paper, the authors have proposed an algorithm for an automatic segmentation of glands in colon histology using local intensity and texture features. Here the dataset images are cropped into patches with different window sizes and taken the intensity of those patches, and also calculated texture-based features. Random forest classifier has been used to classify this patch into different labels. A multilevel random forest technique in a hierarchical way is proposed. This solution is fast, accurate and it is very much applicable in a clinical setup.

## I. INTRODUCTION

Colon Cancer is one of the most common types of cancer, which develops in the colon or rectum (parts of the large intestine) region [1]. It is a kind of tumor, but all tumors are not cancerous. The non-spreadable or benign tumors are not that deadly or destructive as spreadable or malignant one.

Colorectal or colon cancer is the third most common cancer in the world, the third leading cause of cancer-related deaths in women in the United States and the second leading cause in men, with nearly 1.4 million new cases diagnosed in 2012 [2] and about 50,260 deaths during 2017 [3].

Gland Segmentation in Colon Histology images has been done in this paper, using local intensity feature, as well as local texture feature. Local because the patch of the images were taken and the intensity and texture features were calculated of those patches. In this paper, the authors have tried to generate computer program using image processing and machine learning techniques to assist the pathologists for accurate detection of colon cancer [4].

In this project, the authors have used a color normalization technique for colon histology dataset images. After color normalization, the normalized images are cropped into patches with different window sizes corresponding to their Ground truth images provided by the pathologists. Then generates the color histogram of those patches (RGB images) and stores it as feature vector along with level vector which contains three levels Gland, Non-Gland and Mix which are created from the patches of the ground truth images. Before that, ground truth needs to be binarized. In later advancements, texture features and a 2D histogram created after applying a mean filter and a standard deviation filter on the patches were also added as features.

In this project, the Random Forest classifier is used for classification, which randomly grows many classification trees during its training phase. During prediction, each of these trees gives a classification output, and in this manner, each tree cast its 'vote' in favor of a class. The forest chooses that classification which receives most vote (by all the trees in the forest). Random forest runs efficiently on large databases [5].

In the testing phase, it tries to generate or predict the ground truth images from the given RGB images with the help of the trained classifier. It matches the accuracy of the predicted data with the original ground truth given by the pathologists.

All this work is done on the same dataset, which allows the algorithm to be compared, implemented and configured to achieve the optimal performance.

*a) Identification of Need:* Colorectal adenocarcinoma originating in intestinal glandular structures is the most common form of colon cancer. Grading or categorizing them as cancerous or not is still a significant challenge. An automated approach for grading glands is a solution to this problem. This project aims to assist the pathologists in the more accurate detection of colon cancer [6].

*b) Purpose and Scope:* In the world of medicine, pathologists are facing different difficulties in identifying benign and malignant cell quickly and accurately; because of the considerable similarity among them, that job becomes very difficult. Which further leads to wrong treatment and for that, patients and their families have a lot to pay.

In this project, the authors have tried segmenting the glands using Gland segmentation in colon histology images; which helps the pathologist to identify the affected glands using a computer program, which uses image processing and machine learning. This research has the potentiality to assist the pathologist to differentiate between benign and malignant tumor, and it can help the doctors to save the innocent peoples' lives using proper treatments. Reading and analyzeing all the information manually is difficult. So, tools to extract information from large databases are required.

## II. METHODOLOGIES

### A. Flow Diagram

The flow of this algorithm is shown in the Figures 1 and 2, which shows a graphical representation of the algorithm.

### B. Dataset and Parameters

In this project, the dataset of Glass Challenge Contest (GlaS) held at MICCAI'2015[9][10] has been used. The different parameters used for different window or patch size was W1xW1 (W1=21), W2xW2 (W2=11), W3xW3 (W3=5). Further, mean filter as well standard deviation filter of the patches been calculated. To perform the calculation, M1xM1 sliding window size (M1=7) for W1xW1, M2xM2 sliding window size (M2=5) for W2xw2 and M3xM3 (M3=3) for W3xW3 patches has been used.

### C. Proposed Solution

In this project, Random Forest is used to classify the dataset. First, a training dataset with labels are provided to train the dataset.

Like any other supervised learning algorithms, this proposed algorithm takes the input dataset for training, extracts feature, in this case, that would be an RGB Histogram, 2d histogram after applying mean and standard deviation filters and haralick texture features, and feed into the machine learning algorithm, in this case, that would be Random forest.

*a) Training:* The input trained image first Normalized by Reinhard color normalization technique because color variation in image samples is a considerable obstacle to generate reliable results while segmenting images using quantitive analysis. The problem of color constancy in histology images is because they are based on light microscopy and due to variable chemical coloring/reactivity from different manufacturers/batches effects the images differently and also the coloring being dependent on staining procedure (timing, concentrations etc.) [7][8]. To overcome this kind of problem, a Stain Normalization Toolbox is needed. Currently, the toolbox which is used, contains the implementation of Reinhard, RGB Histogram Specification, Macenko, Khan. Among all the available options, Reinhard Normalization was chosen because this method works well for stain dominates of the area of that stain, but it can also provide an undesirable effect on the other stain like turning nuclei black. After color normalization, the normalized images were cropped into different window sizes.

*b) Windows size of patches:* The window size is nothing but the size based on which the RGB and Ground truth images are divided into different patches, in one word it is the size of the patches..

In the next step, the authors have binarized the ground truth images. Image binarization the given image with 256 gray level values to just 0 and 1, so it converts a grayscale image to a black and white image. The easiest approach would be to specify a threshold value, all above that value will be classified as white, rest as black. But this leads to a problem of choosing the threshold value correctly. The best approach would be to use an adaptive binarization approach, which means to have different threshold values for different parts of the image. For this algorithm, a threshold based binarization has be chosen, all the pure black pixel was kept as black and rest was converted to white.

In this proposed algorithm, an RGB histogram, Haralick texture feature, and a 2D histogram after applying the mean filter and standard deviation filter have been taken together as the feature vector. Histogram of red, green and blue channel together as the feature vector and stored these vector in Fr, Fg, Fb matrix accordingly, because in this project the authors got the best result in the combination of all of three. However, anyone can choose any of the combinations of these three channels, like - histogram of Red and Green channel or Green and Blue channel or any one of the channels among this three.

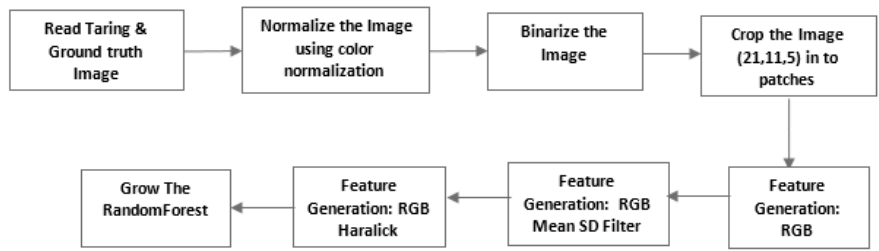

Fig. 1. Control Flow Diagram of the Training Phase

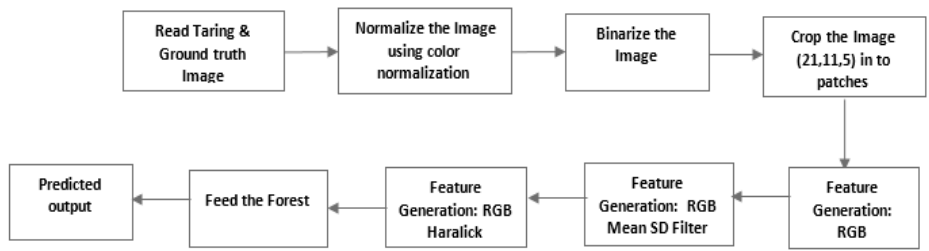

Fig. 2. Control Flow Diagram of the Prediction Phase

Further in the proposed algorithm, the Mean filter as well standard deviation filter were applied on the patches. To perform this operation, M1xM1, M2xM2, M3xM3 sliding window sizes were taken respectively for W1xW1, W2xW2, W3xW3 window sized patches. Then together of the output after applying a mean filter and standard deviation filter, a 2D histogram for three different color channel RGB were created and stored this 2D histogram in FvHist2DImgR, FvHist2DImgG, FvHist2DImgB feature vectors respectively.

Then in the very next step, haralick texture features were calculated and used it as a feature vector along with the RGB histogram and 2D Histogram. In this project for a better result, the authors have used all the 14 features of haralick texture for three different color channel RGB and stored this calculated features in FvHaraR, FvHaraG, FvHaraB features vector correspondingly.

. Just like any other supervised machine learning algorithm, another vector other than feature vector called label vector was created, where Labels of each of the item in the feature vector were stored, in this case for each patch. Label vectors contain various class labels which will be used to train the classifier, and during the classification phase, the classifier will predict any of these class labels.

Next, for classification, the authors have used the Random forest as the classifier. The main reason behind choosing the random forest is that it runs efficiently on large databases. Later on, the performances of other machine learning algorithms were also compared to the random forest.

*c) Prediction:* - In the prediction phase, the goal is to predict the correct label for a given Test image. For doing so, first, the given test image is Normalized just as done in the training phase. Then, the ground truth is binarized. RGB Histogram of those patches was calculated, mean and standard deviation filter were applied, and a 2D histogram was created after applying them, also calculated all the features of the haralick texture of the cropped patches, just as done in training phase. Then using those features, labels were predicted for each patch, and at the end glands segmented output was generated.

*D. Algorithm*

   *a) Training:*

1. Normalize the image using Reinhard Normalization technique
2. Binarize the corresponding ground truth image
3. Divide both normalized image and binarized ground truth image into 21x21 window sized patches
4. Calculate histogram of red, green and blue (RGB) channels of the normalized image patch and store them into Fr, Fg and Fb vectors
5. Apply Mean Filter and Standard Deviation Filter, then create a 2D histogram, separately for RGB channels of the normalized image patches and store them into FvHist2DR, FvHist2DG and FvHist2DB vectors
6. Calculate Haralick Features of the normalized image patches, separately for the RGB channels of the normalized image patch and store them into FvHaraR, FvHaraG and FvHaraB vectors

7. If the highest and the lowest intensity value of the binarized ground truth patch both are equals to 1, mark this patch as 'Gland' in the Label vector, if both are equals to 0, mark this patch as 'non-Gland'. If both doesn't match, mark this patch as 'Mix'
8. Grow a random forest using Fr, Fg, Fb, FvHist2DR, FvHist2DG, FvHist2DB, FvHaraR, FvHaraG and FvHaraB together as feature vector and label vector with tree size 100.
9. Divide 'Mix' patches further into 11x11 window sized patches and repeat step 4 to 8 for those patches.
10. Divide 'Mix' patches further into 5x5 window sized patches and repeat step 4 and 7.
11. For each patch marked as 'Mix', calculate number of white and black pixel in each of them. If number of white pixel is more than number of black pixel, replace the current value in the Label vector with 'Gland', or else 'non-Gland'.
12. Grow a random forest using Fr, Fg, Fb, FvHist2DR, FvHist2DG, FvHist2DB, FvHaraR, FvHaraG and FvHaraB together as feature vector and label vector with tree size 100.

   *b) Prediction:*
1. Normalize the image using Reinhard Normalization technique.
2. Divide normalized image into 21x21 window sized patches.
3. Calculate histogram of RGB channels of the normalized image patch and store them into Fr, Fg and Fb vectors.
4. Apply Mean Filter and Standard Deviation Filter, then create a 2D histogram, separately for RGB channels of the normalized image patches and store them into FvHist2DR, FvHist2DG and FvHist2DB vectors
5. Calculate Haralick Features of the normalized image patches, separately for the RGB channels of the normalized image patch and store them into FvHaraR, FvHaraG and FvHaraB vectors
6. Predict using 1st random forest.
7. Those which are predicted as Mix by the 1st random forest, for those patches, re-patch them with window size 11x11 and predict using 2nd random forest. Choose the most number of times predicted label by the 2nd random forest as the new predicted label for the original 21x21 patch.
8. Those which are predicted as Mix by the 2nd random forest, for those patches, re-patch them with window size 5x5 and predict using 3rd random forest. Choose the most number of times predicted label by the 3rd random forest as the new predicted label for the original 21x21 patch.
9. In the output image, co-ordinates where it's predicted as 'Gland', set them to white with respective window size based on which random forest predicted that.

*E. Post-processing*

The output of the program, even though quite accurate, they do not resemble much with a gland, mainly because of its rigid edges. This post-processing step helps to smooth the edges of the glands, improves the overall structure, as the original algorithm's output has a lot of rigid edges.

   Algorithm:

1. Take one of the output image.
2. Detect edges using Canny Edge Detection.

3. Dilute the detected edges using any of the Morphological structuring element, like disk, diamond, octagon, sphere etc. During testing, it was found out octagon produces best results
4. Add the diluted edges with the original output image to obtain the final output image.

### III. RESULTS & DISCUSSION

#### A. Dataset and Enviornment

Testing of this project was performed using the dataset provided by GlaS@MICCAI'2015: Gland Segmentation Challenge Contest by Department of Computer Science, University of Warwick [9][10]. This dataset already divided into training and testing. Various subsets of the training and testing images were used in combination with various subsets and variations of the algorithm. Testing was done in a system with AMD FX-8350 Octa-core processor (Clock Speed - 4.00 GHz and with Turbo 4.20 GHz) and 32 GB Ram.

#### B. Training and Testing using various Subsets

The first set of training was conducted using a 6-image training dataset, which contains 3 malignant and 3 benign images. Second Set of training was conducted using a 20-image training dataset. Third and final set of training was conducted using all the 85 training images of the GlaS dataset. For testing, the same size of sets were used, but different images.

Various types of classifiers were also created and tested - Single random forest, Hierarchical Random Forest (various window sizes), also random forests with a different number of trees like 10, 100.

Tests were performed with and without color Normalize technique. In this work, Reinhard color normalization technique was used.

*a) Single vs Hierarchical Forest:* While comparing single forest with windows size 40x40 with hierarchical forest with window sizes 40x40, 20x20 and 10x10, both with 100 trees, it was ob-served in some cases single forest works better, but in most cases hierarchical forest performs significantly well over single forest, which is shown in Table I.

*b) With vs Without Color Normalization:* Hierarchical forest with window size 40x40, 20x20 and 10x10and tree size 10 was trained and tested, once without using color normalization and then with using color normali-zation, comparison shown in Table II. It was observed that accuracy was improved in a significant level after color normalization, but in a few rare cases accuracy may decrease.

*c) Number of Trees in Random Forest 10 vs 100:* Hierarchical forest with window size 40x40, 20x20 and 10x10with number of trees 10 and 100 were tested. In most cases, the accuracy increased with increase in the number of trees, but in some cases accuracy decreased as well, shown in Table III.

TABLE I. ACCURACY COMPARISON BETWEEN SINGLE FOREST AND HIERARCHICAL FOREST

| Image | grade (GlaS) | Single Forest | Hierarchical Forest |
|---|---|---|---|
| testA_1 | benign | 0.675 | 0.7214 |
| testA_10 | benign | 0.6821 | 0.55 |
| testA_11 | benign | 0.6607 | 0.5321 |
| testA_12 | malignant | 0.5857 | 0.7429 |
| testA_13 | malignant | 0.6214 | 0.7857 |
| testA_16 | malignant | 0.5389 | 0.7667 |

TABLE II. ACCURACY COMPARISON BETWEEN HIERARCHICAL FOREST WITH AND WITHOUT NORMALIZATION

| Image | | Hierarchical Forest | |
|---|---|---|---|
| | grade (GlaS) | Without Normalization | With Normalization |
| testA_1 | benign | 0.6321 | 0.7214 |
| testA_10 | benign | 0.5179 | 0.55 |
| testA_11 | benign | 0.65 | 0.5321 |
| testA_12 | malignant | 0.7321 | 0.7429 |
| testA_13 | malignant | 0.5357 | 0.7857 |
| testA_16 | malignant | 0.6778 | 0.7667 |

TABLE III. ACCURACY COMPARISON BETWEEN NUMBER OF TREES (TREE SIZE) 10 AND 100

| Image | | Hierarchical Forest | |
|---|---|---|---|
| | grade (GlaS) | Tree Size 10 | Tree Size 100 |
| testA_1 | benign | 0.7214 | 0.6964 |
| testA_10 | benign | 0.55 | 0.5321 |
| testA_11 | benign | 0.5321 | 0.6214 |
| testA_13 | malignant | 0.7857 | 0.8071 |
| testA_16 | malignant | 0.7667 | 0.8111 |

#### C. Training and Testing using the complete Dataset

*a) Priliminary Tests using various window sizes:* There were three distinct combinations were tested for various window sizes. Window sizes 40x40, 20x20 and 10x10 with filter window sizes (for mean and standard deviation filter) 7x7, 5x5 and 3x3, then window sizes 20x20, 10x10 and 5x5 with filter window sizes 7x7, 5x5 and 3x3 and at the end a four-level hierarchical random forest was also tested with window sizes 40x40, 20x20, 10x10 and 5x5 with filter window sizes 7x7, 5x5, 3x3 and 1x1. Performance comparisons is shown in Table IV.

TABLE IV. ACCURACY OF VARIOUS TESTED IMAGES USING VARIOUS WINDOW SIZES

| Window Sizes: - | | 40,20,10 | 20,10,5 | 40,20,10,5 |
|---|---|---|---|---|
| Window Sizes for Filters: - | | 7,5,3 | 7,5,3 | 7,5,3,1 |
| Image | grade (GlaS) | | | |
| testA_1 | benign | 0.378571 | 0.528965 | 0.392857 |
| testA_10 | benign | 0.582143 | 0.636277 | 0.535714 |
| testA_11 | benign | 0.675 | 0.660969 | 0.664286 |
| testA_12 | malignant | 0.753571 | 0.755935 | 0.764286 |
| testA_13 | malignant | 0.807143 | 0.82906 | 0.825 |
| testA_14 | malignant | 0.625 | 0.68661 | 0.642857 |
| testA_15 | malignant | 0.378571 | 0.37037 | 0.389286 |
| …… | …… | …… | ……. | …… |
| testA_40 | benign | 0.646429 | 0.773029 | 0.6 |
| testA_41 | malignant | 0.771429 | 0.777778 | 0.764286 |
| testA_42 | malignant | 0.703571 | 0.689459 | 0.671429 |
| …… | ……. | ……. | ……. | ……. |
| testB_1 | malignant | 0.807143 | 0.790123 | 0.785714 |
| testB_10 | malignant | 0.635714 | 0.65812 | 0.639286 |
| testB_11 | malignant | 0.732143 | 0.768281 | 0.732143 |
| testB_12 | malignant | 0.682143 | 0.730294 | 0.692857 |
| ….. | …… | ……. | ……. | …… |
| testB_9 | malignant | 0.710714 | 0.736942 | 0.696429 |
| *Average* | | *0.677730325* | *0.702068288* | *0.664852* |

It has been observed that window size 20,10,5 produces best results. So, finally window sizes 21,11,5 was

choosen – near odd numbers of 20,10,5 for the final rounds of testing.

*b) Final rounds of tests:* After various tests, it was found that 3 level hierarchical classifier using windows sizes 21, 11 and 5 has proven to be the best in terms of accuracy. As random forest is a statistical model, every time the model is retrained and tested produces different accuracy. Though most of the times these accuracies are not that different, but sometimes they can be widely varied. To test the algorithm in a better manner, the same model using the same algorithm is trained and tested 5 times to check performance consistancy. The complete comparison is shown in Table VI. Also, 5 round average is shown.

*c) Comparison of Random Forest with other Models, using same algorithm:* The proposed algirthm uses Random Forest as it's Machine Learning model. But for comparing its accuracy supoiririty with other famous machine learing models such as k-Nearest Neighbour (KNN) and Support Vector Machine (SVM) they were tested as well. Same window sizes been used for all the tests. Results are displayed in Table V. It has been observed that Random Forest is superior to both KNN and SVM.

### D. Some output of the program

Some outputs of the program which are generated after training and testing on the whole dataset using window sizes 21,11,5 and window sizes for filters 7, 5, 3. Results of the last round is shown in Figure 3 to 6. After the output is obtained from the model, those output images were again processed using the proposed post processing algorithm which is mentioned in II.E. This post-processing step helps to smooth the edges of the glands, improves the overall structure, as the original algorithm's output has a lot of rigid edges.

TABLE V. COMPARISON OF RANDOM FOREST (AVERAGE OF ALL 5 ROUNDS) WITH KNN AND SVM

| Window Sizes: - | | 21,11,5 | | |
|---|---|---|---|---|
| Window Sizes for Filters: - | | 7,5,3 | | |
| Image | grade (GlaS) | Random Forest | KNN | SVM |
| testA_1 | benign | 0.5552434 | 0.522162 | 0.313514 |
| testA_10 | benign | 0.6086486 | 0.587027 | 0.438919 |
| testA_11 | benign | 0.6406486 | 0.345946 | 0.643243 |
| testA_12 | malignant | 0.8038918 | 0.645405 | 0.718919 |
| testA_13 | malignant | 0.8467024 | 0.747027 | 0.779459 |
| testA_14 | malignant | 0.7072432 | 0.664865 | 0.721081 |
| testA_15 | malignant | 0.3963246 | 0.513514 | 0.571892 |
| ……. | ……. | ……. | ……. | …… |
| testA_26 | malignant | 0.7364326 | 0.691892 | 0.743784 |
| testA_27 | benign | 0.6931892 | 0.647568 | 0.478919 |
| testA_28 | benign | 0.569946 | 0.578378 | 0.381622 |
| testA_29 | malignant | 0.7844328 | 0.682162 | 0.739459 |
| ……. | ……. | ……. | ……. | ……. |
| testA_38 | malignant | 0.7249732 | 0.691892 | 0.714595 |
| testA_39 | malignant | 0.8015134 | 0.713514 | 0.712432 |
| ……. | ……. | ……. | ……. | ……. |
| *Average* | | *0.70436379* | *0.638549* | *0.60097* |

## IV. FUTURE SCOPE AND APPLICATIONS

In this project, authors have tried segmenting the glands using Gland segmentation in colon histology images; which is being the first step towards cancer detection. After the program correctly identifies the glands, then another program can be written which will be able to differentiate those detected glands as malignant or benign.. This research has the potentiality to assist the pathologist to differentiate between benign and malignant tumor, and it can help the doctors to save the innocent peoples' lives using proper treatment classification. If any other set of histology images and corresponding ground truth images are provided to this algorithm for training, this algorithm can segment glands in those pictures as well. This algorithm can also be adapted for segmenting other pictures as well, with necessary changes.

TABLE VI. SAME ALGOIRTHM USED IN 5 ROUNDS, ALSO AVERAGE OF ALL THE ROUNDS

| Window Sizes: - | | 21,11,5 | | | | | |
|---|---|---|---|---|---|---|---|
| Window Sizes for Filters: - | | 7,5,3 | | | | | |
| Image | grade (GlaS) | Round 1 | Round 2 | Round 3 | Round 4 | Round 5 | Average |
| testA_1 | benign | 0.555676 | 0.561081 | 0.548108 | 0.555676 | 0.555676 | *0.555243* |
| testA_10 | benign | 0.610811 | 0.606486 | 0.604324 | 0.610811 | 0.610811 | *0.608649* |
| testA_11 | benign | 0.64 | 0.643243 | 0.64 | 0.64 | 0.64 | *0.640649* |
| testA_12 | malignant | 0.8 | 0.812973 | 0.806486 | 0.8 | 0.8 | *0.803892* |
| testA_13 | malignant | 0.845405 | 0.856216 | 0.841081 | 0.845405 | 0.845405 | *0.846702* |
| testA_14 | malignant | 0.708108 | 0.705946 | 0.705946 | 0.708108 | 0.708108 | *0.707243* |
| testA_15 | malignant | 0.394595 | 0.397838 | 0.4 | 0.394595 | 0.394595 | *0.396325* |
| testA_16 | malignant | 0.792208 | 0.788961 | 0.808442 | 0.792208 | 0.792208 | *0.794805* |
| testA_17 | malignant | 0.592432 | 0.605405 | 0.597838 | 0.592432 | 0.592432 | *0.596108* |
| …… | …… | ……. | ……. | …… | …… | …… | …… |
| testB_1 | malignant | 0.787027 | 0.802162 | 0.792432 | 0.787027 | 0.787027 | *0.791135* |
| testB_10 | malignant | 0.700541 | 0.690811 | 0.695135 | 0.700541 | 0.700541 | *0.697514* |
| testB_11 | malignant | 0.763243 | 0.762162 | 0.756757 | 0.763243 | 0.763243 | *0.76173* |
| testB_12 | malignant | 0.748108 | 0.740541 | 0.748108 | 0.748108 | 0.748108 | *0.746595* |
| testB_13 | malignant | 0.722162 | 0.724324 | 0.707027 | 0.722162 | 0.722162 | *0.719567* |
| testB_14 | malignant | 0.687568 | 0.703784 | 0.721081 | 0.687568 | 0.687568 | *0.697514* |
| …… | …… | ….. | …… | ……. | …… | …… | …… |
| testB_7 | benign | 0.846486 | 0.843243 | 0.84973 | 0.846486 | 0.846486 | *0.846486* |
| testB_8 | malignant | 0.764324 | 0.775135 | 0.772973 | 0.764324 | 0.764324 | *0.768216* |
| testB_9 | malignant | 0.755676 | 0.742703 | 0.752432 | 0.755676 | 0.755676 | *0.752433* |
| *Average* | | *0.71325* | *0.713876* | *0.7134452* | *0.713275* | *0.713275* | *0.71343* |

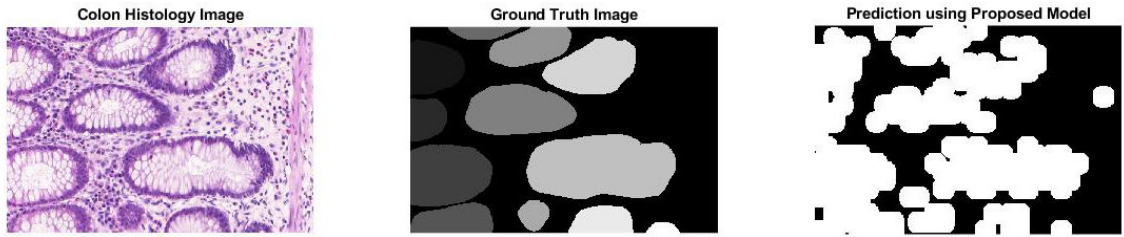

Fig. 3. testA_52 (Average accuracy of 5 rounds: 84%)

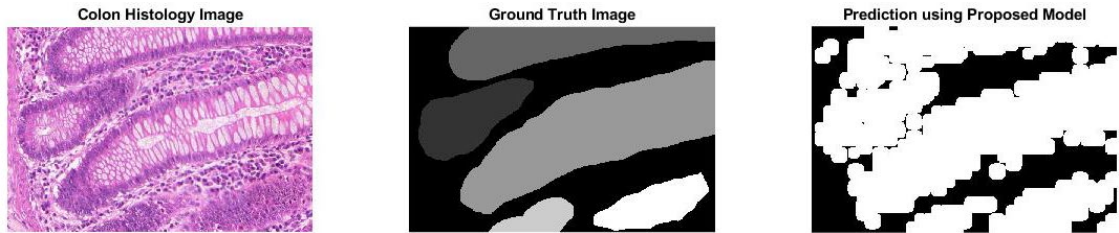

Fig. 4. testB_7 (Average accuracy of 5 rounds: 84%)

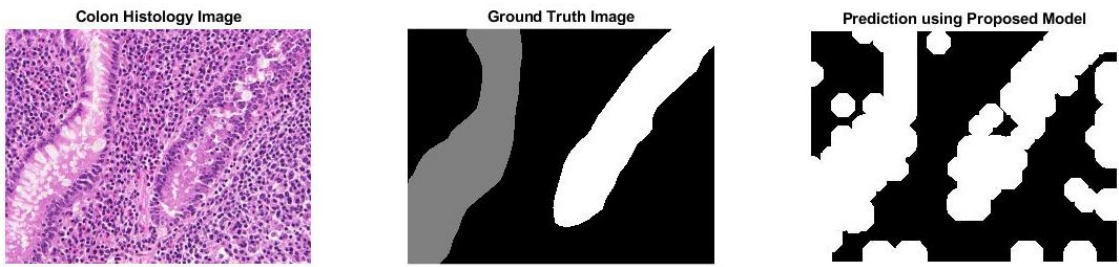

Fig. 5. testA_2 (Average accuracy of 5 rounds: 80%)

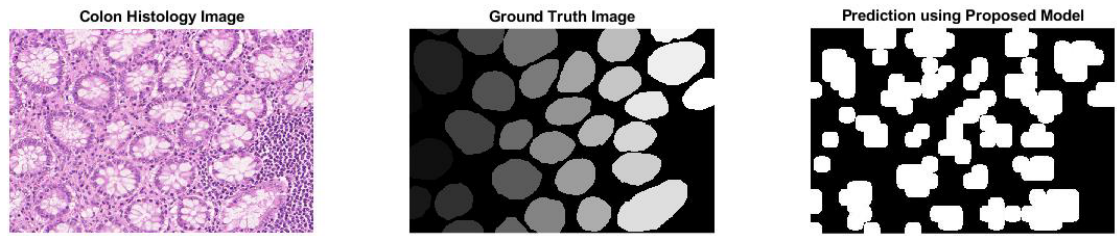

Fig. 6. testA_27 (Average accuracy of 5 rounds: 69%)

## V. Conclusion

The goal of this work was to develop a robust and efficient system, using the available tools and algorithms, which can be an elegant system that segments glandular regions in colon histology images. This algorithm is versatile enough to be used with other histology images, by changing the dataset.

In this work, the authors have presented a novel model for segmenting glandular structures in histology images of human colon tissues. The authors have used Gland Segmentation in Colon Histology Images and used the color histogram, mean filter, standard deviation filter, haralick texture feature and 2d histogram as feature vectors. Random forest has been used as the classifier in this project.

The principal objective of this paper is to assist the pathologist to accurate detection of colon cancer.

Using this program, pathologists will just have to give the input image, and it will produce the desired outcome with segmented glands

## Acknowledgment

This work was made possible by the Electronics and Communication Sciences Unit (ECSU), Indian Statistical Institute (ISI), Kolkata in association with the Indian Unit for Pattern Recognition and Artificial Intelligence (IUPRAI) at ISI.

The authors would like to thank Dr. Angshuman Paul (Senior Research Fellow, Electronics and Communication Sciences Unit, Indian Statistical Institute, Kolkata) for his guidance to understand the subject, his encouragement, his help in every step and Dr. Abhishek Roy (Former H.O.D, Amity Institute of Information Technology (AIIT), Amity University, Kolkata) for his motivation and support.